\documentclass[10pt, a4paper]{article}
\usepackage{lrec}
\usepackage{multibib}
\newcites{languageresource}{Language Resources}
\usepackage{graphicx}
\usepackage{tabularx}
\usepackage{soul}

\usepackage{epstopdf}
\usepackage[utf8]{inputenc}

\usepackage{hyperref}
\usepackage{xstring}
\usepackage{color}
\usepackage{amsmath}
\usepackage{url}
\usepackage{float}
\usepackage{booktabs}
\usepackage{subcaption}
\usepackage{microtype}
\usepackage[
  separate-uncertainty = true,
  multi-part-units = repeat
]{siunitx}
\usepackage{multicol}
\usepackage{multirow}
\usepackage{array}
\usepackage{tabu,colortbl}
\usepackage{hhline}

\makeatletter
\newcommand*{\rom}[1]{\expandafter\@slowromancap\romannumeral #1@}
\makeatother
\usepackage[T1]{fontenc}
\usepackage{csquotes}

\usepackage{enumitem}

\usepackage{bm}

\newcommand{\ex}[1]{\textsl{#1}} 

\title{Decorate the Examples:\\ A Simple Method of Prompt Design for Biomedical Relation Extraction
}

\name{Hui-Syuan Yeh, Thomas Lavergne, Pierre Zweigenbaum}

\address{Université Paris-Saclay, CNRS, Laboratoire Interdisciplinaire des Sciences du Numérique, Orsay, France \\
         \{yeh, lavergne, pz\}@lisn.fr\\}

\abstract{
Relation extraction is a core problem for natural language processing in the biomedical domain. Recent research on relation extraction showed that prompt-based learning improves the performance on both fine-tuning on full training set and few-shot training. However, less effort has been made on domain-specific tasks where good prompt design can be even harder. In this paper, we investigate prompting for biomedical relation extraction, with experiments on the ChemProt dataset. We present a simple yet effective method to systematically generate comprehensive prompts that reformulate the relation extraction task as a cloze-test task under a simple prompt formulation. 
In particular, we experiment with different ranking scores for prompt selection. With BioMed-RoBERTa-base, our results show that prompting-based fine-tuning obtains gains by 14.21 F1 over its regular fine-tuning baseline, and 1.14 F1 over SciFive-Large, the current state-of-the-art on ChemProt. Besides, we find prompt-based learning requires fewer training examples to make reasonable predictions. The results demonstrate the potential of our methods in such a domain-specific relation extraction task.}

\begin{document}

\maketitleabstract

\section{Introduction}
With the rapid growth of biomedical textual resources in scientific articles, clinical notes, patient forums, social media, and so on, helping humans quickly grasp the key information out of vast content
has become necessary.
Natural Language Processing and more specifically Information Extraction (IE) algorithms support readers by transforming unstructured text into structured information of interest. Relation extraction (RE), as one of the most important IE tasks, focuses on recognizing the relation types between two entities mentioned in a given sentence (e.g., given \ex{Alfred Hitchcock directed Psycho}, identify that the relation between (\ex{Alfred Hitchcock}, \ex{Psycho}) is \ex{DirectorOf}). 

The current state of the art in information extraction is obtained by Transformer models such as BERT \cite{Devlin-NAACLHLT2019}.  Great success has been obtained by
adapting BERT architectures to biomedical tasks by additional training (BioBERT \cite{lee2020biobert}, ClinicalBERT \cite{alsentzer-etal-2019-clinicalbert}), or by pretraining from scratch (SciBERT \cite{beltagy-etal-2019-scibert}, PubMedBERT \cite{gu2020pubmedbert}) on biomedical text corpora. 
More recently, BioMegatron \cite{shin-etal-2020-biomegatron}
studied the pretraining settings better for the biomedical BERT models; CharacterBERT \cite{el-boukkouri-etal-2020-characterbert} enabled
word representations without requiring segmentation into a priori word pieces,
to better represent domain-specific terms in specialized domains. Another stream works on incorporating external knowledge bases into models \cite{michalopoulos-etal-2021-umlsbert,liu2021sapbert}.

Compared to previous work that augments training data with biomedical textual or structured data, we explore an alternative training paradigm, \emph{prompting}, to adapt pre-trained models to biomedical RE tasks more efficiently. 
The current dominant paradigm consists in pre-training a neural model with a language modeling objective such as masked word prediction \cite{Devlin-NAACLHLT2019}, then fine-tuning this model by retraining it with a different objective related to the target task (e.g., relation extraction).
The main idea of prompting, on the other hand, is to keep the language modeling objective as it is, so that pre-trained models can be put to use more directly and efficiently to address the downstream task.

Compared to the other line of works on low resource \cite{sample-acl,curriculum-eacl,select-acl}, prompting boosts the performance in the few-shot setting without picking or sorting the training examples. In general, prompting has been shown to be efficient in recent work for a number of downstream tasks \cite{NEURIPS2020-brown,schick-etal-2020-automatically}. 
 Its benefits for domain-specific relation extraction have however received less attention.

The contributions of our paper are as follows:
\begin{itemize}
    \item We explore prompting on biomedical relation extraction with the ChemProt dataset.
    \item We present a systematic approach for prompt design in relation extraction tasks for a specific domain without manual effort, including a variety of ranking scores for prompt selection.
    \item Our results show that prompting boosts model performance,
      both when fine-tuning on the full training set and in a few-shot training condition.
\end{itemize}

\section{Background}

BERT \cite{Devlin-NAACLHLT2019}, RoBERTa \cite{liu2019roberta}, and other pre-trained models revolutionized the IE field with universal model designs that are capable of fitting almost all linguistic tasks with minimum change.  These models can adapt from pre-training to fine-tuning on various downstream tasks. Thus, the dominant approach for IE tasks nowadays is to adapt these pre-trained language models via objective engineering.  However, we can alleviate the gap between the two phases even further by reformulating the fine-tuning tasks into the form of the pre-training task, i.e.\ masked word prediction. This training paradigm, known as \emph{prompting}, has been proven to be efficient in adapting to downstream tasks in prior work. We refer interested readers to a recent systematic survey on prompting studies \cite{liu2021survey} for more detail. 

The main idea of prompting is to reformulate the given tasks into \emph{templates} with blank positions (e.g., \ex{Steve Jobs left Apple in 1985. Steve Jobs is the \underline{\hspace{0.5cm}} of Apple}) and ask a language model to score how well \emph{label words}, i.e., words associated with relations labels, fill these blanks (e.g., \ex{founder}). The majority of earlier work uses only one word to fill the blank, though it is often difficult to accommodate more complicated relation with one label word. (e.g., Relation: \ex{place\_of\_birth}, with the example \ex{Juan Laporte (born November $24$, $1959$) is a former boxer who was born in Guayama, Puerto Rico.}) On top of that, say we work with binary relations, e.g.
\begin{itemize}
    \item Relation: \ex{founders}, with example like: \ex{Steve Jobs left Apple in 1985.} 
    \item Relation: \ex{nationality}, with example like: \ex{Aragaki Yui is an Japanese actress.}
\end{itemize}

Applied with the previous template, \ex{Aragaki Yui is the \underline{\hspace{0.5cm}} of Japanese} would not make sense for the newly introduced relation, \ex{nationality}. In practice, most relation extraction tasks are multi-class classification problems which makes the design of template and the corresponding label words even harder.

Coming up with good templates and label words is the key to good performance. Recent work on prompt-based learning contributes various template schemes, for instance for fact probing \cite{petroni-etal-2019-language,jiang-etal-2020-x}, text classification \cite{gao2021lm-bff}, question answering \cite{khashabi-etal-2020-unifiedqa,jiang-etal-2021-qa}, or commonsense reasoning \cite{DBLP:journals/corr/abs-1806-02847} in the general domain.

\paragraph{Prompting in Relation Extraction}
Relation extraction (RE) is a multi-class classification problem that involves classifying the relation between two entities within or across sentences. It is non-trivial to manually design appropriate prompts to distinguish many different classes. Close to \cite{schick-schutze-2021formulation} experimenting with different prompt templates for binary and ternary classification tasks, \newcite{chen2021adaptprompt} introduced an interpretable and intuitive template for RE to alleviate the required manual effort in a large search space, specifically leaving the label words for human design.  An example is \ex{[E1] Google [/E1] is [MASK] [E2] Alphabet [/E2]}, where \ex{[MASK]} represents a blank.  Under this template formulation, \newcite{han2021ptr} added extra blanks to fill before the two entities for incorporating entity type information (e.g., \ex{the [MASK] Google [MASK][MASK][MASK] the [MASK] Alphabet}, is expected to be filled as: the \ex{organization} Google \ex{’s parent was} the \ex{organization} Alphabet) and \newcite{chen2021knowprompt} presented a synergistic optimization over entity types and relation labels that results in virtual label words. 

On another note, \newcite{shin2020auto} performed gradient-guided search to automatically search for a suitable template and label words for each class. The resulting prompts 
are often uninterpretable and inconsistent across relations, hence hard for the models to work with, although their prompt generation is fully automatic. In our work, we start from the template format of \cite{han2021ptr} and extend it to a systematic generation of comprehensive prompts without human effort.

\paragraph{Prompting for Biomedical Information Extraction}
\newcite{sung-etal-2021-language} released BioLAMA, a benchmark composed of biomedical
factual knowledge triples for probing biomedical language models. They showed that biomedical language models yield better predictions compared to general models, but they also found that it is due to the model predictions being biased towards certain prompts. To help applying language models with prompt-based learning, \newcite{2021rarewords} proposed a method to paraphrase rare words with the help of an extra source (Wiktionary\footnote{https://www.wiktionary.org/}) for natural language inference (NLI) and Semantic Textual Similarity (STS) tasks in the clinical domain.

Prompt-based few-shot learning and fine-tuning have gained attention in the general domain, but is still under-explored in specialized domains. In this paper, we investigate prompting for relation extraction in the biomedical domain.

\begin{figure*}[htbp] 
\centering
\resizebox{1\textwidth}{!}{\includegraphics{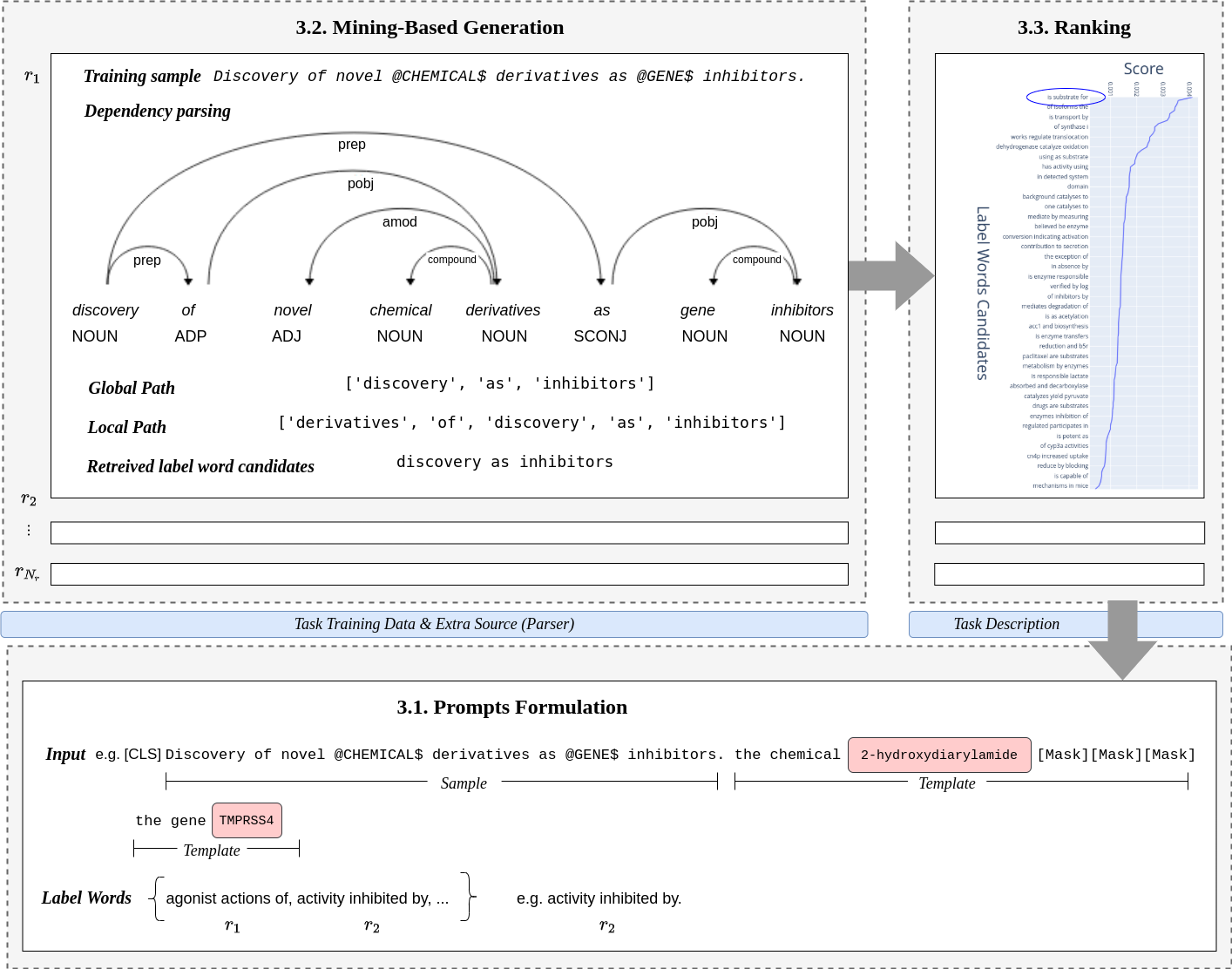}}
\caption{An illustration of the method. Blue marks the resources we use for prompt engineering, red marks the entities.}
\label{fig:method}
\end{figure*}

\section{Method}
Relation extraction involves identifying the relation type between two entities.  We address intra-sentence relations, which are the most frequent in most datasets. For ease of discussion, we will refer to the two entities as $e_{1}$ and $e_{2}$, which in our case are a chemical and gene respectively. We begin by introducing the prompt formulation in \ref{section:3.1}, which describes how the examples fed to the models are decorated. Then, we unfold how we come up with the components required for completing the formulation, by first collecting the candidates for the components in \ref{section:3.2}, and selecting candidates with proposed ranking scores in \ref{section:3.3}. Figure \ref{fig:method} illustrates our method.

\subsection{Prompt Formulation}
\label{section:3.1}
We illustrate below how we prepare examples for the relation extraction task conventionally (1) and with prompting (2). 

\begin{enumerate}[label*=(\arabic*)]
    \item 
    (Input) \ex{The specificity of tracer uptake was determined by adding the [E1] imipramine [/E1] inhibitor [E2] NET [/E2].} 
    \item[] (Label) CPR:4 
    \item 
    (Input) \ex{The specificity of tracer uptake was determined by adding the \emph{imipramine} inhibitor \emph{NET}. imipramine \underline{\hspace{1cm}} NET.}
    \item[] (Label Words) \ex{is inhibitor of}
\end{enumerate}

Following the simple template proposed by \cite{han2021ptr}, we reformulate each example by appending to it a sentence containing its two entities, with masked tokens between them. In this prompting setting,
label words must be defined
for each relation. We make room for multiple masked words for better expressiveness, and choose a fixed number of 3 words for simplicity. The model is then expected to score sequences of label words for every relation. The key to model performance lies in choosing relevant label words depending on the task. 

\subsection{Mining-based Label Word Generation}
\label{section:3.2}

\newcite{toutanova-etal-2015-representing} pointed out that sentences containing synonymous textual relations often share common words, sub-structure, and have similar syntactic dependency arcs. \newcite{jiang2020know} followed that line and used words on the shortest dependency paths between the two entities as label words.  This method however often retrieves label words found around the entities rather than between them and hence does not fit our template formulation. Instead, we identify the \emph{local path}: the shortest dependency path from $e_{1}$ to $e_{2}$ and the \emph{global path} the shortest path from the first word to the last word of a sentence.\footnote{We use  the spaCy dependency parser, https://spacy.io/api/dependencyparser} We take the words appearing on both the \emph{global path} and the \emph{local path} and prune the rest of the words.

\subsection{Ranking}
\label{section:3.3}
To choose the most relevant label words among those mined for each relation $r$, we score the label word candidates $c$ based upon how salient the word is for the relation. We discuss ranking scores $R(c, r)$ based upon different features. In our notation, $N_{c}(r)$ is the number of examples labelled $r$ in which candidate $c$ occurs, $N_{r}$ is the number of relations $r$ in which candidate $c$ occurs, $N_{R}$ is the total number of relations, $\tilde{c}$ and $\tilde{r}$ are the sentence embeddings for $c$ and for the description of  relation $r$.

\paragraph{Frequency} This score directly obtains clues from the training set by checking the number of occurrences. 
\begin{align}
	R_{\textrm{frequency}}(c, r) &= N_{c}(r).
	\label{eq:frequency}
\end{align}

\paragraph{Frequency-Specificity} The principle is close to tf.idf which suggests that label words that are shared across all relation types are not relevant. This score is defined as:
\begin{align}
	R_{\textrm{frequency-specificity}}(c, r) &= N_{c}(r) \log 
	\frac{N_{R}}{N_{r}}.
	\label{eq:frequency-specificity}
\end{align}

\paragraph{Similarity} This score attempts to take the task description into consideration. The frequency score might select irrelevant words that are far from the meaning of the relation type. We use here the cosine similarity between the sentence embeddings of the candidate words and of the relation description:
\begin{align}
	R_{\textrm{similarity}}(c, r) &= cos(\tilde{c}, \tilde{r}).
	\label{eq:similarity}
\end{align}

\paragraph{Combined} This score combines the above statistical and semantic properties 
and is calculated as follows:
\begin{align}
	R_{\textrm{combined}}(c, r) &= R_{\textrm{frequency-specificity}}(c, r) \cdot R_{\textrm{similarity}}(c, r).
	\label{eq:combined}
\end{align}

\section{Experiments} 
\subsection{Dataset}
We use the ChemProt dataset \citelanguageresource{kringelum2016chemprot} to investigate the relation extraction task. It contains scientific paper abstracts annotated with 6 relation types between chemicals and genes in sentences: activation (CPR:3), inhibition (CPR:4), agonist (CPR:5), antagonist (CPR:6), substrate (CPR:9), and no relation. The details are presented in Table~\ref{tab:dataset}. 

\begin{table}[H]
\centering
\resizebox{0.75\columnwidth}{!}{
	\begin{tabular}{crrr}
	\toprule
	\textbf{Relation} & \textbf{N. train} & \textbf{N. val} & \textbf{N. test} \\
        \midrule
        All & 19,460 & 11,820 & 16,943 \\ 
        \midrule
        CPR:3 & 2,251 & 1,094 & 1,661 \\ 
        CPR:4 & 768 & 550 & 665  \\ 
        CPR:5 & 727 & 457 & 644  \\ 
        CPR:6 & 235 & 199 & 293  \\ 
        CPR:9 & 173 & 116 & 195  \\ 
        No Relation & 15,306 & 9,404 & 13,485  \\ 
        \bottomrule
	\end{tabular}
	}
  	\caption{Description of the ChemProt dataset.}
  	\label{tab:dataset}
\end{table}

\subsection{Model}
We conduct experiments with the off-the-shelf Roberta-base\footnote{https://huggingface.co/roberta-base} and BioMed-RoBERTa-base \footnote{https://huggingface.co/allenai/biomed\_roberta\_base} pre-trained language models. BioMed-RoBERTa-base is continuously pre-trained on scientific biomedical articles based on the RoBERTa-base architecture. Both models have obtained good performance on biomedical domain tasks \cite{liu2019roberta,biomed-roberta} including the relation extraction task we are studying. For the baselines, we add a linear layer on top of the final hidden state of the [CLS] token to pull out the predictions.  For the prompting method, we take the outputs of the masked positions from the last hidden layer, then calculate the similarities with the label word embeddings: these similarity scores serve as our model predictions.

\subsection{Hyperparameter  Settings}
We train with 5 epochs with batch size 8. The AdamW optimizer is used with a learning rate of 3e-5, weight decay rate 1e-2, and epsilon 1e-6.
For fine-tuning on the whole training split, we report results over 5 random initializations. For few-shot experiments, the performance of learning with few steps can vary significantly depending on the choice of training and validation splits. To mitigate this instability, performance results are averaged over 5 runs on different random seeds to split into training and validation splits. Specifically, $k$ stands for the number of examples we draw from each relation type; we resample from the pool for the few cases of a relation containing fewer examples than $k$.
Because of the unbalanced distribution of the dataset, some earlier work applies re-sampling, weighting, or simply exclude the dominant class (\emph{no relation}). On the contrary, we do not employ any extra strategy to reshape the distribution, and examine whether the models can cope with it on their own. 

\subsection{Experiments}
Under the RoBERTa architectures, we set up experiments to compare the prompt-based learning and the regular supervised learning without prompts, i.e., we add a sequence classification head on top of the pre-trained language models and perform fine-tuning, on both general and biomedical models. In addition, within the prompt-based learning, we set up experiments for different ranking metrics and their counterpart, random pick without any ranking. Lastly, we evaluate on few-shot settings on RoBERTa-base, where we take prompt-based learning with the ranking metric $R_{\textrm{combined}}(c, r)$, which is the best for fine-tuning on the full training set, and regular supervised learning.

\section{Results}

\subsection{Results on Prompting}
Table \ref{tab:performance} shows results for fine-tuning on the full training set. Overall, we see that prompting indeed boosts up the performance for both models, especially with BioMed-RoBERTa-base achieving the best results $90.09$ (sd: $0.08$). To the best of our knowledge, the best result so far was $88.95$, accomplished by SciFive-Large \cite{phan2021scifive}, a heavier model based on T5 \cite{2020t5}, which we outperform in the present work. 

We experiment with label words selected with the proposed ranking scores and as well as a random pick from the candidate pool without ranking. The results show that ranking scores do help, especially RoBERTa-base performs best with prompts generated with $R_{\textrm{combined}}(c, r)$ and BioMed-RoBERTa-base performs best with $R_{\textrm{frequency}}(c, r)$. We expected that $R_{\textrm{combined}}(c, r)$ would be the best ranking scores; however, BioMed-RoBERTa-base might carry some knowledge on the biomedical vocabulary, causing similarity and specificity not to contribute much and frequency to obtain the top results. Note that our F1-score for BioMed-RoBERTa-base without prompting is behind that reported in the source ($83.0$, sd: $0.7$). This might be due to the different hyperparameter setting and to the relation class weighting. 

\begin{table}[htbp]
\centering
\resizebox{1\columnwidth}{!}{
	\begin{tabular}{clccc}
	\toprule
	\textbf{Model} & \textbf{Ranking} & \textbf{Micro F1} & \textbf{Macro F1} \\
        \midrule
        RoBERTa-base & - &  80.09 (0.12) & 19.23 (0.63) \\ 
        BioMed-RoBERTa-base & - & 76.69 (0.10) & 17.20 (0.91) \\
        \midrule
        RoBERTa-base & random pick &  88.17 (0.28) & 72.08 (0.50) \\ 
         & frequency &  88.12 (0.51) & 72.26 (0.60) \\ 
         & frequency-specificity & 88.35 (0.11) & 72.38 (0.68) \\ 
         & similarity & 88.43 (0.38) & 73.02 (0.80) \\ 
         & combined & 88.60 (0.13) & 74.13 (3.06)  \\ 
         \rule{0pt}{3ex} 
        BioMed-RoBERTa-base & random pick &  89.55 (0.14) & 74.79 (0.41) \\ 
        & frequency & \textbf{90.09 (0.08)} & \textbf{76.31 (0.23)}  \\ 
        & frequency-specificity & 90.09 (0.15) & 76.17 (0.19) \\
        & similarity & 89.99 (0.15) & 75.64 (0.50) \\
        & combined & 89.78 (0.33) & 75.65 (0.70) \\
        \bottomrule
	\end{tabular}
	}
  	\caption{Performance (\%) on prompt fine-tuning with prompts generated by different ranking scores and conventional fine-tuning. We report the average and standard deviation over 5 random runs.}
  	\label{tab:performance}
\end{table}

\subsection{Few-Shot Learning on Prompting}
\begin{figure}[ht]
\centering
\resizebox{\columnwidth}{!}{\includegraphics{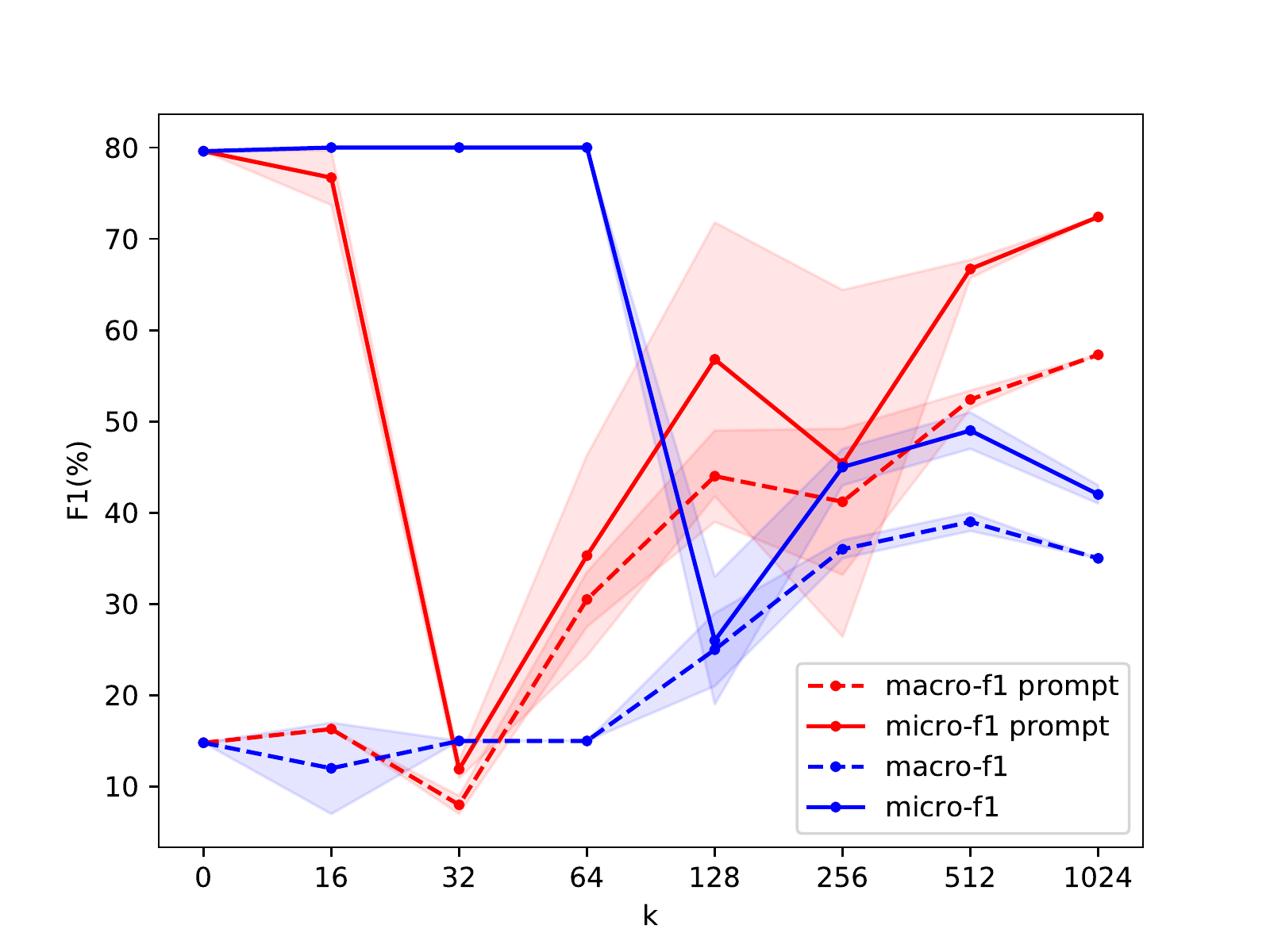}}
\caption{Performance (\%) over few-shot examples.}
\label{few-shot}
\end{figure}

In Figure \ref{few-shot}, we show our few-shot learning experiments with Roberta-base. We use the ranking metric $R_{\textrm{combined}}(c, r)$ for prompting. Both approaches start with a high micro-f1 score, but low macro-f1: the predictions are both all on the majority class. We see that for prompting, a dramatic drop of performance occurs at $k=32$, which for the baseline occurs later at $k=128$. This drop is a turning point where the models start to learn meaningful predictions instead of always predicting the major relation type.
Having this turning point earlier shows the better behavior of the prompting method.

\section{Conclusion} 
In this paper, we investigate prompting for biomedical relation extraction. We propose methods to systematically generate comprehensive prompts to reformulate a relation extraction task. Under a simple prompt template, label word candidates are mined from the training set with the help of a parser, and we propose various ranking metrics to select the best label words representing the relations. Our results show that prompting outperforms the de~facto training paradigm to apply pre-trained models. The results demonstrate the potential of our methods for domain-specific relation extraction tasks. To advance further, there are still many future directions and possible improvements for the approach: (1) as the label words candidate pool can be small, augmenting the pool with knowledge bases and other existing resources, (2) aggregating multiple label words, and (3) mitigating the bias that language models have with label word calibration \cite{zhao2021calibrate}.

\section{Acknowledgements}
This research was supported in part by project KEEPHA, ANR-20-IADJ-0005-01, under the trilateral ANR-DFG-JST AI call.

\section{Bibliographical References}
\bibliographystyle{lrec} 
\bibliography{literature}

\section{Language Resource References} 
\bibliographystylelanguageresource{lrec}
\bibliographylanguageresource{dataset}

\section*{Appendix} 

\begin{table*}[!htbp]
    \begin{center}
    \resizebox{\textwidth}{!}{%
    \begin{tabular}{cccccc}
      \hline
      \\[-1em]
      Relations & Random Pick & Frequency & Frequency-Specificity & Similarity & Combined \\ \\[-1em]
      \hline \\[-1em]
      CPR:3 & of src stimulates & is activated by & is activated by & is activated by & is activated by \\
      
      CPR:4 & was difference between & is inhibitor of & design as inhibitors & of inhibition by & activity inhibited by \\
      CPR:5 & are gene the & activity is mediated & activity is mediated & agonist actions of & agonist actions of \\
      CPR:6 & features of receptor & identified are antagonists & identified are antagonists & known as antagonist & identified are antagonists \\
      CPR:9 & was greater in & involved in secretion & involved in secretion & is substrate for & is substrate for \\
      No Relation & effect evaluated in & by concentrations of & by concentrations of & was unable bind & by concentrations of \\
      
      \hline \\[-1em]
    \end{tabular}
}
\caption{Extracted label words with various ranking metrics}
 \end{center}
\end{table*}

\end{document}